%% file: acl2021.tex
\documentclass[11pt,a4paper]{article}
\usepackage[hyperref]{acl2021}
\usepackage{times}
\usepackage{latexsym}

\usepackage{graphicx}
\usepackage{caption}
\usepackage{stfloats}
\usepackage{latexsym}
\usepackage{booktabs}
\usepackage{CJKutf8}
\usepackage{color}
\usepackage{subfiles}
\usepackage{multirow}
\usepackage{amsmath}
\usepackage[ruled]{algorithm2e}
\usepackage{enumitem}
\usepackage{bbold}

\usepackage{subfigure}

\usepackage{microtype}

\aclfinalcopy 
\def\aclpaperid{376} 

\title{A Multimodal Sentiment Dataset for Video Recommendation}
\author{
Hongxuan Tang, \ \ \ 
Hao Liu, \ \ \  
Xinyan Xiao, \ \ \ 
Hua Wu \\
Baidu Inc., Beijing, China\\
 \{{\tt tanghongxuan, liuhao24, xiaoxinyan, wu\_hua\}@Baidu.com}
}

\date{}

\begin{document}
\maketitle

\begin{abstract}

\subfile{sections/Abstract}

\end{abstract}

\section{Introduction}

\subfile{sections/Introduction}

\section{Related Work}

\subfile{sections/RelatedWork}

\section{Data Construction}

\subfile{sections/Dataset}

\section{Experiments}
\subfile{sections/Experiments}

\section{Conclusion}

\subfile{sections/Conclusion}


\bibliographystyle{acl_natbib}
\bibliography{anthology,acl2021}


\end{document}

%% file: sections/Abstract.tex
Recently, multimodal sentiment analysis has seen remarkable advance and a lot of datasets are proposed for its development. In general, current multimodal sentiment analysis datasets usually follow the traditional system of sentiment/emotion, such as positive, negative and so on. However, when applied in the scenario of video recommendation, the traditional sentiment/emotion system is hard to be leveraged to represent different contents of videos in the perspective of visual senses and language understanding. Based on this, we propose a multimodal sentiment analysis dataset, named baiDu Video Sentiment dataset (DuVideoSenti), and introduce a new sentiment system which is designed to describe the sentimental style of a video on recommendation scenery. Specifically, DuVideoSenti consists of 5,630 videos which displayed on Baidu \footnote{One of the most popular applications in China, which features both information retrieval and news recommendation}, each video is manually annotated with a sentimental style label which describes the user's real feeling of a video. Furthermore, we propose UNIMO \cite{li2020unimo} as our baseline for DuVideoSenti. Experimental results show that DuVideoSenti brings new challenges to multimodal sentiment analysis, and could be used as a new benchmark for evaluating apporaches designed for video understanding and multimodal fusion. We also expect our proposed DuVideoSenti could further improve the development of multimodal sentiment analysis and its application to video recommendations.

%% file: sections/Introduction.tex

Sentiment analysis is an important research area in Natural Language Processing (NLP), which has wide applications, such as opinion mining, dialogue generation and recommendation. Previous work \cite{liu2012survey, tian2020skep} mainly focused on text sentiment analysis and achieved promising results. Recently, with the development of short video applications, multimodal sentiment analysis has obtained more attention \cite{tsai2019multimodal, li2020unimo,  yu2021learning} and a lots of datasets \cite{li2017cheavd, poria2018meld, yu2020ch} are proposed to advance its developments. However, current multimodal sentiment analysis datasets usually follow the traditional sentiment system (positive, neutral and negative ) or emotion system (happy, sad, surprise and so on), which is far from satisfactory especially for video recommendation scenario of application.

In the scenery of industrial video recommendation, content-based recommendation method \cite{pazzani2007content} is widely used because of its advantages in improving the explainability of recommendation and conducting effective online interventions. Specifically, the videos are first represented by certain tags, which is automatically predicted by some neural network models \cite{wu2015fusing, wu2016multi, rehman2021deep}. Then, a user's profile is constructed by gathering the tags from his/her watched videos. Final, it recommends candidate videos based on the relevance between current candidate video's tags and the user's profile. In general, the types of tags are further divided into topic-level and entity-level. As show in Figure 1, for the videos talking about one's traveling in The Palace Museum, they may relate to the topic-level tags such as "Tourism" and entity-level tags such as "The Palace Museum". While in real application, the topic-level and entity-level are not afford to summary the content of a certain video comprehensively. The reason relies in that the videos, which share the same topics or contains similarity entities, usually gives different visual and sentimental perceptions to a certain user. For example, the first video brings us a feeling of great momentum, while the second video brings us a feeling of generic, although they share the same topics.

Based on this, we propose a multimodal sentiment analysis dataset, named baiDu Video Sentiment dataset (DuVideoSenti), and construct a new sentimental -style system, which consists of eight sentimental-style tags. These sentimental-style tags are designed to describe the users' real feeling after he/she has watched the video.

In detail, we collect 5,630 videos which displayed on Baidu, and then each video is manually annotated with a sentimental-style label selected from the above sentimental-style system. In addition, we further propose a baseline to our DuVideoSenti Corpus which is based on UNIMO. And we expect our proposed DuVideoSenti corpus could further improve the development of multimodal sentiment analysis and its application to video recommendations.

\begin{figure*}
\centering
    {
        \includegraphics[width=0.95\columnwidth]{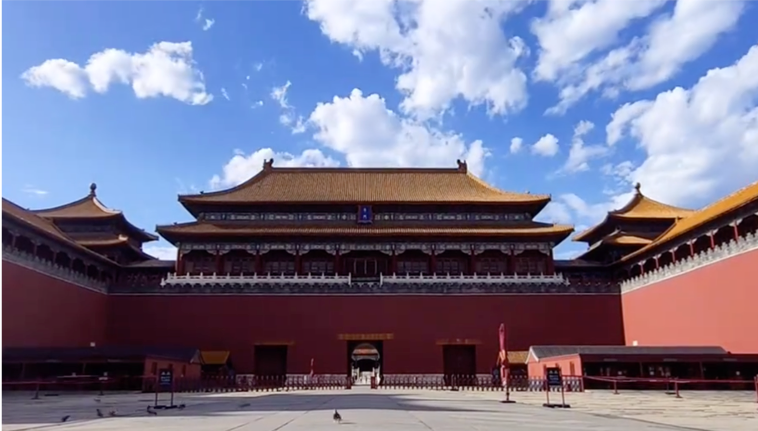} 
    }
    \hspace{.2in}
    {
        \includegraphics[width=0.95\columnwidth]{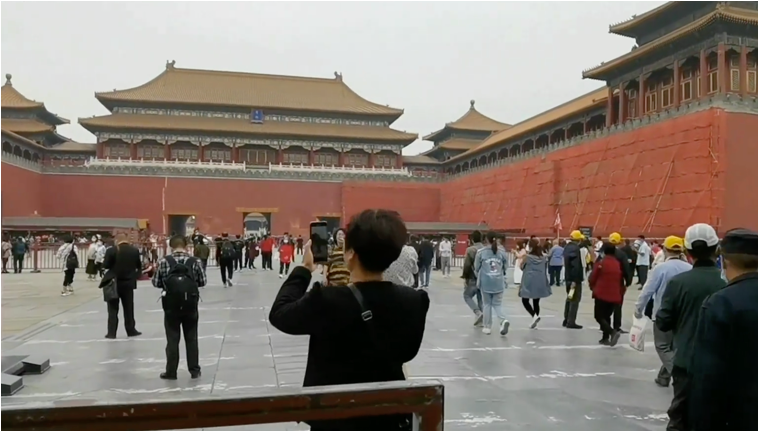} 
    }
\caption{Two videos displayed in Baidu, which record the authors' travelling to The Palace Museum. Although, the two videos share the same topic-level and entity-level tags such as "Tourism" and "The Palace Museum", they give different visual and sentimental perceptions to the audience. Specifically, the first video gives us a feeling of great momentum, while the another one brings us a feeling of generic.}
\label{example-figure}
\end{figure*}


%% file: sections/Relatedwork.tex
In this section, we briefly introduce previous works in multimodal datasets, multimodal sentiment analysis.

\subsection{Multimodal Datasets}

To promote the development of multimodel sentiment analysis and emotion detection, a variety of multimodal datasets are proposed, including IEMOCAP \cite{busso2008iemocap}, YouTube \cite{morency2011towards}, MOUD \cite{perez2013utterance}, ICT-MMMO \cite{wollmer2013youtube}, MOSI \cite{zadeh2016mosi}, CMU-MOSEI \cite{zadeh2018multimodal}, CHEAVD \cite{li2017cheavd}, Meld \cite{poria2018meld}, CH-SIMS \cite{yu2020ch} and so on. As mentioned above, current multimodal datasets follow the traditional sentiment/emotion system, which is not fit to the scenario of video recommendations. In this paper, our DuVideoSenti dataset defines a new sentimental-style system which is customized for video recommendation.

\subsection{Multimodal Sentiment Analysis}

In multimodal sentiment analysis, intra-modal representation and inter-modal fusion are two essential and challenging subtasks. For intra-modal representation, previous work pay attention to the temporal and spatial characteristics among different modalities. The Convolutional Neural Network (CNN), Long Short-term Memory (LSTM) and Deep Neural Network (DNN) are representative approaches to extract multimodal features \cite{cambria2017benchmarking, zadeh2017tensor, zadeh2018memory}. For inter-modal fusion, numerous methods have been proposed recently. For example, concatenation \cite{cambria2017benchmarking}, Tensor Fusion Network (TFN) \cite{zadeh2017tensor}, Low-rank Multimodal Fusion (LMF) \cite{liu2018efficient}, Memory Fusion Network (MFN) \cite{zadeh2018memory}, Dynamic Fusion Graph (DFG) \cite{zadeh2018multimodal}. Recently, we follow the trend of pretaining, and select UNIMO \cite{li2020unimo} to build our baselines.

%% file: sections/Dataset.tex

In this section, we introduce the constrcution details of our DuVideoSenti dataset. Specifically, DuVideoSenti contains 5,630 video examples, all examples are selected from Baidu. Besides, in order to ensure the data quality, all sentiment labels are annotated by experts. Finally, we release DuVideoSenti according to following data structure as shown in Table~\ref{dataset-struct}. Table 1 shows a real example in our DuVideoSenti datasets, which contains multi-regions such as url, title, video feaures and its corresponding sentiment label. For the video features extraction, we first sample four frame images from all the frame images of the given video at the same time intervals. Second, we use Faster R-CNN \cite{ren2015faster} to detect the salient image regions and extract the visual features (pooled ROI features) for each region, which is the same as \cite{chen2020uniter,li2020unimo}. Specifically, the image-region features are set as 100.

\begin{table*}[tbp]
\centering
\begin{tabular}{ll}
\toprule
\textbf{Region} & \textbf{Example} \\ 
\midrule
url & http://quanmin.baidu.com/sv?source=share-h5\&pd=qm\_share\_search\& \\
& vid=5093910907173814607
\\
title & \begin{CJK*}{UTF8}{gbsn}那小孩不让我吃口香糖…\end{CJK*} \\
label & \begin{CJK*}{UTF8}{gbsn}呆萌可爱\end{CJK*}  \\

\multirow{7}*{frame1} & \multirow{7}*{\begin{minipage}[b]{1.2\columnwidth}
		\centering
		\raisebox{-.9\height}{\includegraphics[width=\linewidth]{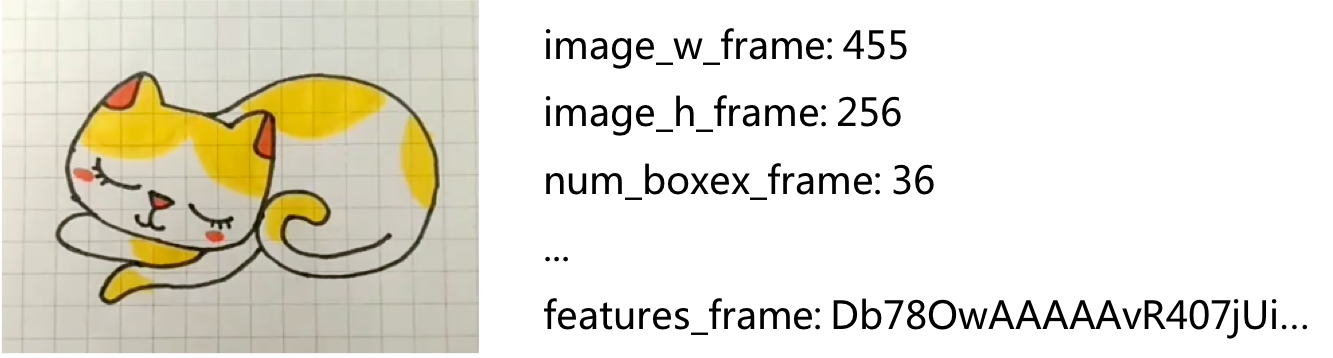}}
	\end{minipage} 
}  \\
&  \\
& \\
& \\
& \\
& \\
& \\

... & ... \\

\multirow{7}*{frame4} & \multirow{7}*{\begin{minipage}[b]{1.2\columnwidth}
		\centering
		\raisebox{-.9\height}{\includegraphics[width=\linewidth]{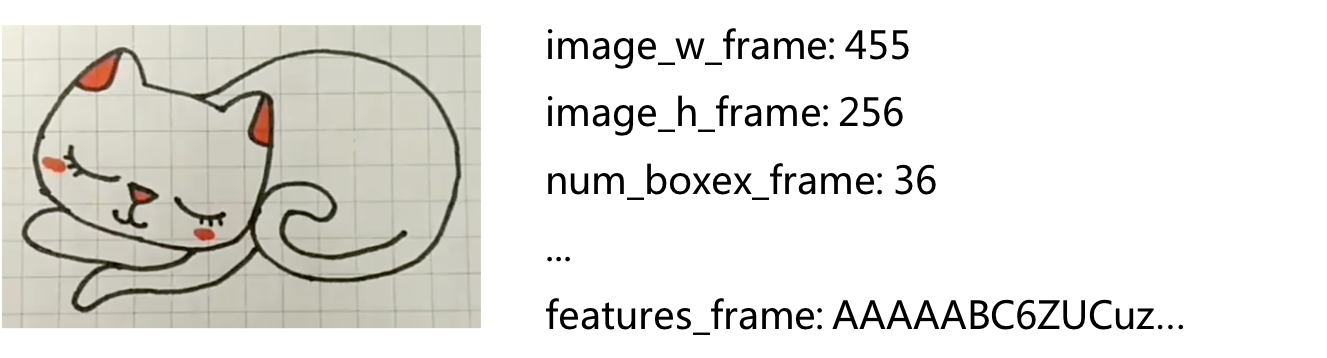}}
	\end{minipage} 
}  \\
&  \\
& \\
& \\
& \\
& \\
& \\

\bottomrule
\end{tabular}
\caption{Data structure of released dataset. (It should be noted that we have not provide the origin frame images in our dataset for the protection of copyright. Instead, we provided the visual features extracted from each frame image and its corresponding url for watching the video online.)}
\label{dataset-struct}
\end{table*}

In addition, we split DuVideoSenti into training set and test set randomly, the size of which are 4,500 and 1,130 respectively. The sentimental-style labels are listed as follows:
\begin{CJK*}{UTF8}{gbsn}
``文艺清新\ (Hipsterism)'', ``时尚炫酷\ (Fashion)'', ``舒适温馨\ (Warm and Sweet)'', ``客观理性\ (Objective and Rationality)'', ``家长里短\ (Daily)'', ``土里土气\ (Old-fashion)'', ``呆萌可爱\ (Cute)'', ``奇葩低俗\ (Vulgar)'', ``正能量\ (Positive Energy)'', ``负能量\ (Negative Energy)'', ``其他\ (Other)''
\end{CJK*}, the class distribution and examples of each sentiment label in our DuVideoSenti are shown in Table~\ref{dataset-example}.

\begin{table*}[h]
\centering
\setlength{\tabcolsep}{4mm}
\begin{tabular}{llc}
\toprule
\textbf{Label} & \textbf{Example} & \textbf{\#} \\ 
\midrule
\begin{CJK*}{UTF8}{gbsn}文艺清新\end{CJK*} &
\begin{CJK*}{UTF8}{gbsn}令箭荷花的开放过程\end{CJK*} & 
\multirow{2}{*}{520} \\
Hipsterism & Translated: The opening process of Nopalxochia  & 
\vspace{1.2mm} \\
\begin{CJK*}{UTF8}{gbsn}时尚炫酷\end{CJK*} & 
\begin{CJK*}{UTF8}{gbsn}古驰中国新年特别系列\end{CJK*} & 
\multirow{2}{*}{193} \\
Fashion & Translated: Gucci Chinese New Year limited edition &
\vspace{1.2mm} \\
\begin{CJK*}{UTF8}{gbsn}舒适温馨\end{CJK*} &
\begin{CJK*}{UTF8}{gbsn}美好的一天，从有爱的简笔画开始\end{CJK*}& 
\multirow{2}{*}{485} \\
Warm and Sweet & Translated: A beautiful day, start with simple strokes of love &
\vspace{1.2mm} \\
\begin{CJK*}{UTF8}{gbsn}客观理性\end{CJK*} & 
\begin{CJK*}{UTF8}{gbsn}吹气球太累?是你没找对方法！\end{CJK*} & 
\multirow{2}{*}{1,204} \\
Objective and Rationality & Translated: Tired of blowing balloons? Try this method. & 
\vspace{1.2mm} \\
\begin{CJK*}{UTF8}{gbsn}家长里短\end{CJK*} &
\begin{CJK*}{UTF8}{gbsn}今天太阳炎热夏天以到来\end{CJK*}  & 
\multirow{2}{*}{1,334} \\
Daily & Translated: It's very hot today, summer is coming & 
\vspace{1.2mm} \\
\begin{CJK*}{UTF8}{gbsn}土里土气\end{CJK*} &
\begin{CJK*}{UTF8}{gbsn}广场舞，梨花飞情人泪32步\end{CJK*} & 
\multirow{2}{*}{71}  \\
Old-fashion & Translated: The 32 steps of public square dancing & 
\vspace{1.2mm} \\
\begin{CJK*}{UTF8}{gbsn}呆萌可爱\end{CJK*} &
\begin{CJK*}{UTF8}{gbsn}\#创意简笔画\#可爱小猫咪怎么画？\end{CJK*}  & 
\multirow{2}{*}{522} \\
Cute &  Translated: \#Creative stick figure\#How to draw a cute kitten?  & 
\vspace{1.2mm} \\
\begin{CJK*}{UTF8}{gbsn}奇葩低俗\end{CJK*} & 
\begin{CJK*}{UTF8}{gbsn}撒网是我的本事，入网是你的荣幸\end{CJK*} & 
\multirow{2}{*}{282} \\
Vulgar & Translated: It is your honour to love me & 
\vspace{1.2mm} \\
\begin{CJK*}{UTF8}{gbsn}正能量\end{CJK*}   &
\begin{CJK*}{UTF8}{gbsn}山东齐鲁医院130名医护人员集体出征\end{CJK*} & 
\multirow{2}{*}{81} \\
Positive Energy & Translated: Shandong Qilu Hospital 130 medical staff set out  & 
\vspace{1.2mm} \\
\begin{CJK*}{UTF8}{gbsn}负能量\end{CJK*}   & 
\begin{CJK*}{UTF8}{gbsn}黑社会被打屁股\end{CJK*}   & 
\multirow{2}{*}{5}  \\
Negative Energy & Translated: The underworld spanked & 
\vspace{1.2mm} \\
\begin{CJK*}{UTF8}{gbsn}其他\end{CJK*}     &
\begin{CJK*}{UTF8}{gbsn}速记英语，真的很快记住\end{CJK*} & 
\multirow{2}{*}{933} \\ 
Other & Translated: English shorthand, really quick to remember & 
\\
\midrule
\textbf{Total}    &  & \textbf{5,630} \\ 
\bottomrule
\end{tabular}
\caption{Distribution and examples for different sentiment classes in DuVideoSenti.}
\label{dataset-example}
\end{table*}

%% file: sections/Experiments.tex

\subsection{Experiment Setting}
In our experiment, we select UNIMO-large , which consists of 24 layers of Transformer block,as as our baseline. The maximum sequence length of text tokens and image-region features are set as 512 and 100, respectively. The learning rate is set to 1e-5 and the batch size is $8$. We set the number of epochs to $20$. All experiments are conducted on 1 Tesla V100 GPUs. We select Accuracy to evaluate the performance of the baseline systems. 

\subsection{Experimental Results}

\begin{table}[h!]
\centering
\begin{tabular}{lccc}
\toprule
  & \textbf{frame 1} & \textbf{frame 4} & \textbf{frame 20} \\ \midrule
\begin{CJK*}{UTF8}{gbsn}文艺清新\end{CJK*} & 50.91 & 52.88 & 45.19 \\
\begin{CJK*}{UTF8}{gbsn}时尚炫酷\end{CJK*} & 2.56  & 28.20 & 15.38 \\
\begin{CJK*}{UTF8}{gbsn}舒适温馨\end{CJK*} & 27.85 & 27.83 & 37.11 \\
\begin{CJK*}{UTF8}{gbsn}客观理性\end{CJK*} & 62.65 & 67.91 & 65.14 \\
\begin{CJK*}{UTF8}{gbsn}家长里短\end{CJK*} & 64.41 & 67.69 & 74.90 \\
\begin{CJK*}{UTF8}{gbsn}土里土气\end{CJK*} & 6.66  & 0.00  & 6.66  \\
\begin{CJK*}{UTF8}{gbsn}呆萌可爱\end{CJK*} & 60.95 & 39.04 & 68.57 \\
\begin{CJK*}{UTF8}{gbsn}奇葩低俗\end{CJK*} & 29.82 & 33.33 & 10.52 \\
\begin{CJK*}{UTF8}{gbsn}正能量\end{CJK*}   & 5.88  & 41.17 & 47.05 \\
\begin{CJK*}{UTF8}{gbsn}负能量\end{CJK*}   & 0.00  & 0.00  & 0.00  \\
\begin{CJK*}{UTF8}{gbsn}其他\end{CJK*}     & 57.75 & 59.89 & 56.14 \\ 
\midrule
\textbf{All}    & \textbf{52.65} & \textbf{54.56} & \textbf{56.46} \\ 
\bottomrule
\end{tabular}
\caption{UNIMO baseline performance based on different number of key frames.}
\label{exp-main-res}
\end{table}

\begin{table}[h!]
\centering
\setlength{\tabcolsep}{4mm}
\begin{tabular}{lcccc}
\toprule
  & \textbf{Visual} & \textbf{Textual} & \textbf{Multi} \\ \midrule
\begin{CJK*}{UTF8}{gbsn}文艺清新\end{CJK*} & 52.88  & 32.69 & 52.88\\
\begin{CJK*}{UTF8}{gbsn}时尚炫酷\end{CJK*} & 17.94  & 17.94 & 28.20\\
\begin{CJK*}{UTF8}{gbsn}舒适温馨\end{CJK*} & 18.55  & 27.83 & 27.83\\
\begin{CJK*}{UTF8}{gbsn}客观理性\end{CJK*} & 66.39  & 56.84 & 67.91\\
\begin{CJK*}{UTF8}{gbsn}家长里短\end{CJK*} & 61.79  & 65.54 & 67.69\\
\begin{CJK*}{UTF8}{gbsn}土里土气\end{CJK*} & 6.66   & 0.00  & 0.00\\
\begin{CJK*}{UTF8}{gbsn}呆萌可爱\end{CJK*} & 49.52  & 49.52 & 39.04\\
\begin{CJK*}{UTF8}{gbsn}奇葩低俗\end{CJK*} & 35.08  & 28.07 & 33.33\\
\begin{CJK*}{UTF8}{gbsn}正能量\end{CJK*}   & 11.76  & 47.05 & 41.17\\
\begin{CJK*}{UTF8}{gbsn}负能量\end{CJK*}   & 0.00   & 0.00  & 0.00\\
\begin{CJK*}{UTF8}{gbsn}其他\end{CJK*}     & 56.68  & 41.71 & 59.89\\ 
\midrule
\textbf{All}    & \textbf{51.85} & \textbf{47.25} & \textbf{54.56}\\ 
\bottomrule
\end{tabular}
\caption{Baseline performance based on image and text.}
\label{exp-img-text}
\end{table}

Table 3 shows the baseline performance in our dataset, specifically, UNIMO-large model achieves an Accuracy of 54.56\% on the test set. It suggests that the by simply using a strong and widely used models failed to obtain promising results, which proposes the needs for further development of multimodal sentiemnt analysis, especially for videos. 

We are also interested in the affects brought by different number of image-region frames. We further test the performance of our baseline system when $\{1, 4, 20\}$ frames are selected. The experiment results are listed in Table~\ref{exp-main-res}, which shows that the performance of decreases with less image-region frames. It suggests that by improving visual representation, can further promote the classification performance.



Finally, we propose to evaluate the improvements brought by multi-modal fusion. We compare the accuracy performance among the systems which use visual, textual, and multi-modal information respectively. The results are show in Table~\ref{exp-img-text}, which shows that by fusing the visual and textual information of the videos, it obtained the best performance. Furthermore, visual-only models performs better than the textual-only one, we suspect that is our defined sentimental-style system is much more related to user' visual feelings after he/she watched the video.

%% file: sections/Conclusion.tex
In this paper, we propose a new multimodel sentiment analysis dataset named DuVideoSenti, which is designed for the scenario of video recommendation. Furthermore, we propose UNIMO as our baseline, and test the accuray performance of the baseline on a variety of settings. We expect our DuVideoSenti dataset could further improve the development for the area of multimodal sentiment analysis, and promote the applications to video recommendation.